\pdfoutput=1

\documentclass[11pt]{article}

\usepackage[]{acl}
\usepackage{times}
\usepackage{latexsym}
\usepackage{graphicx}
\usepackage[T1]{fontenc}

\usepackage[utf8]{inputenc}

\usepackage{microtype}
\usepackage{hhline}
\usepackage{xspace}
\usepackage{soul}
\usepackage{booktabs}
\usepackage{enumitem}
\usepackage{algorithm}
\usepackage{algpseudocode}
\usepackage{multirow}
\usepackage{amsmath}

\definecolor{brightmaroon}{rgb}{0.76, 0.13, 0.28}

\newcommand{\sys}{\textsc{Primera}\xspace}
\newcommand{\pegasus}{\textsc{Pegasus}\xspace}
\newcommand{\pretraining}{pretraining\xspace}

\newcommand{\pretrained}{pretrained\xspace}
\newcommand{\pretrain}{pretrain\xspace}

\algnewcommand\algorithmicrequireb{{\hspace{0.85cm}}}
\algnewcommand\INPTDESCB{\item[\algorithmicrequireb]}

\algnewcommand\algorithmicfuncdesc{\textbf{Function:}}
\algnewcommand\FUNCDESC{\item[\algorithmicfuncdesc]}
\algnewcommand\algorithmicfuncdescb{{\hspace{1.48cm}}}
\algnewcommand\FUNCDESCB{\item[\algorithmicfuncdescb]}
\algnewcommand{\algorithmicgoto}{\textbf{goto}}
\algnewcommand{\Goto}[1]{\algorithmicgoto~\ref{#1}}

\title{\sys: Pyramid-based Masked Sentence Pre-training for \\ Multi-document Summarization }

\author{Wen Xiao\footnotemark[2]\hspace{.3em}\thanks{~~Work mainly done during an internship at AI2.}\qquad Iz Beltagy\footnotemark[3]\hspace{1.9em} Giuseppe Carenini\footnotemark[2]\hspace{1.9em} Arman Cohan\footnotemark[3]\hspace{.35em}\footnotemark[4]\vspace{4pt}\\
\footnotemark[2]\hspace{.4em}University of British Columbia, Vancouver, Canada 
\\ \footnotemark[3]\hspace{.4em}Allen Institute for AI, Seattle, WA, USA \\ 
\footnotemark[4]\hspace{.4em}Paul G. Allen School of Computer Science \& Engineering, University of Washington\\
\texttt{\small{\{xiaowen3,carenini\}@cs.ubc.ca}},
\texttt{\small{\{beltagy,armanc\}@allenai.org
}}}

\begin{document}
\maketitle
\begin{abstract}

We introduce \sys, a pre-trained model for multi-document 
representation with a focus on summarization that reduces the need for dataset-specific architectures and large amounts of fine-tuning labeled data. 
\sys uses our newly proposed pre-training objective designed to teach the model to connect and aggregate information across documents. 
It also uses efficient encoder-decoder transformers to simplify the processing of concatenated input documents.
With extensive experiments on 6 multi-document summarization datasets from 3 different domains on zero-shot, few-shot and full-supervised settings, \sys outperforms 
current state-of-the-art dataset-specific and pre-trained models on most of these settings with large margins.\footnote{The code and pre-trained models can be found at \url{https://github.com/allenai/PRIMER}}
\end{abstract}

\section{Introduction}
Multi-Document Summarization is the task of generating a summary from a cluster
of related documents.
State-of-the-art approaches to multi-document summarization are primarily either graph-based~\cite{liao-etal-2018-abstract,li-etal-2020-leveraging-graph,pasunuru-etal-2021-efficiently}, leveraging graph neural networks to connect information between the documents, or hierarchical ~\cite{liu-lapata-2019-hierarchical,fabbri-etal-2019-multi,jin-etal-2020-multi}, building intermediate representations of individual documents and then aggregating information across. While effective, these models either require domain-specific additional information e.g. Abstract Meaning Representation~\cite{liao-etal-2018-abstract}, or discourse graphs~\cite{christensen-etal-2013-towards,li-etal-2020-leveraging-graph},
or use dataset-specific, customized architectures, making it difficult to leverage \pretrained language models. Simultaneously, recent \pretrained language models (typically encoder-decoder transformers) have shown the advantages of \pretraining and transfer learning for generation and summarization \cite{t5,lewis-etal-2020-bart,longformer,bigbird}.
Yet, existing \pretrained models either use single-document \pretraining objectives
or use encoder-only models that do not work for generation tasks like 
summarization ~\citep[e.g., CDLM, ][]{cdlm}.

\begin{figure}[t]
    \centering
    \includegraphics[width=0.9\linewidth]{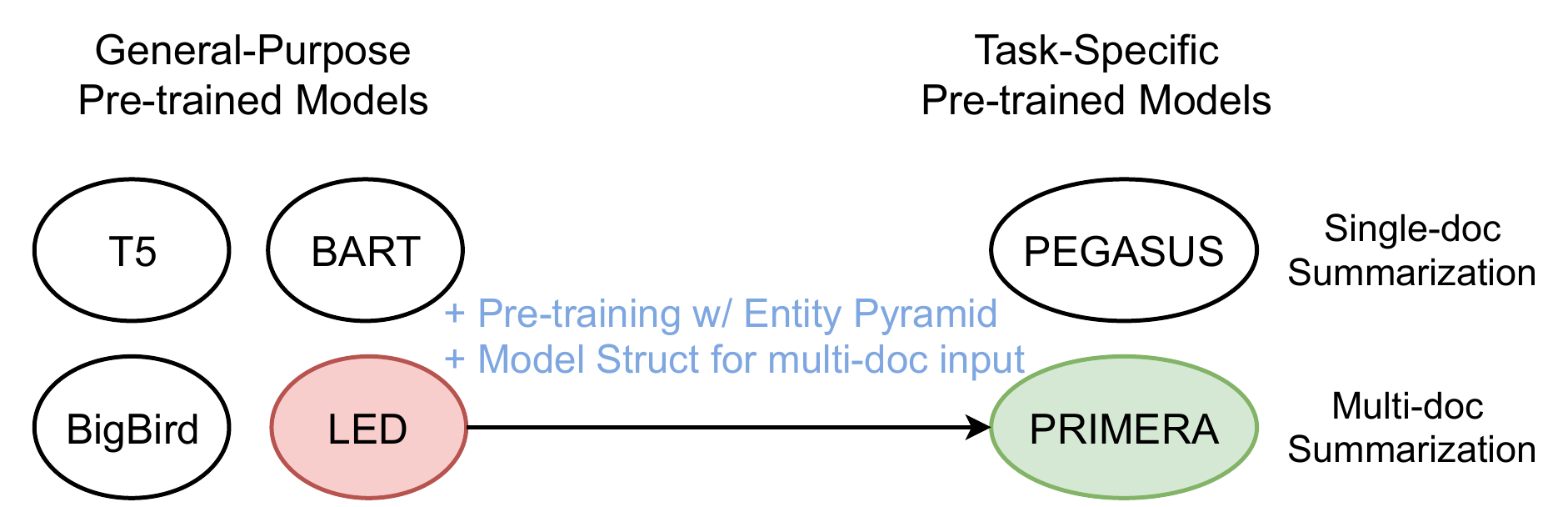}
    \caption{
    \sys vs existing pretrained models. 
    }
    \label{fig:intro}
\end{figure}

\begin{figure*}[t]
    \centering
    \includegraphics[width=1\linewidth]{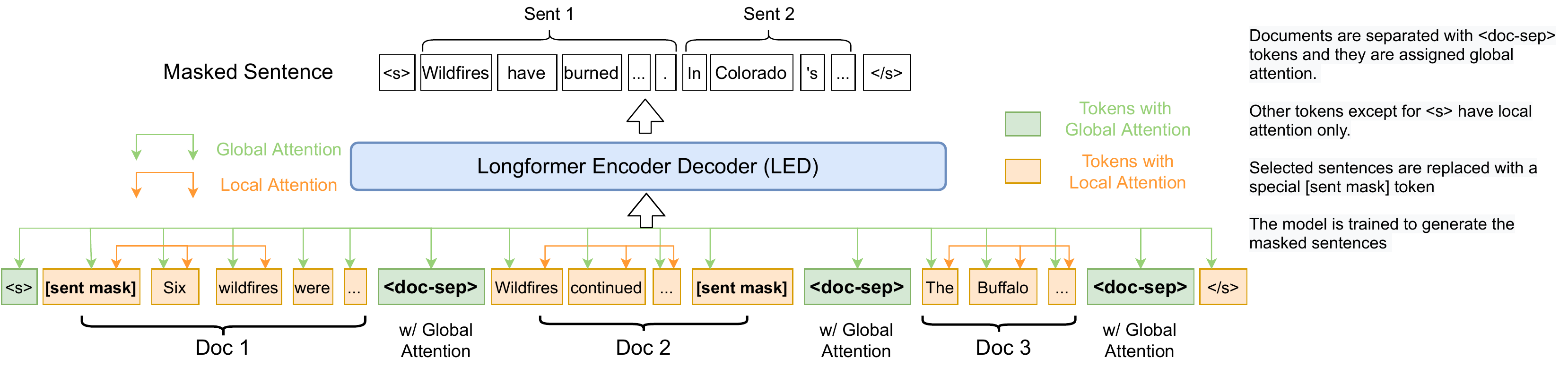}
    \caption{Model Structure of \sys.
    }
    \label{fig:model_structure}
\end{figure*}

Therefore, we argue that these \pretrained models are not necessarily the best fit for multi-document summarization. 
Alternatively, we propose a simple
\pretraining approach for multi-document summarization, reducing the need for dataset-specific architectures and large fine-tuning labeled data (See Figure \ref{fig:intro} to compare with other \pretrained models). 
Our method is designed to teach the model to identify and aggregate salient information across a ``cluster'' of related documents during \pretraining.
Specifically, our approach uses the Gap Sentence Generation objective (GSG)~\cite{pegasus}, i.e. masking out several sentences from the input document, and recovering them in order in the decoder. 
We propose a novel strategy for GSG sentence masking which we call, Entity Pyramid, inspired by the Pyramid Evaluation method~\cite{nenkova-passonneau-2004-evaluating}.
With Entity Pyramid, we mask salient sentences in the entire cluster 
then train the model to generate them, encouraging it to find important information across documents and aggregate it in one summary.

We conduct extensive experiments on \textbf{6} multi-document summarization datasets from \textbf{3} different domains. We show that despite its simplicity, \sys achieves superior performance compared with prior state-of-the-art \pretrained models, as well as dataset-specific models in both few-shot and full fine-tuning settings. \sys performs particularly strong in zero- and few-shot settings, significantly outperforming prior state-of-the-art up to \textbf{5} Rouge-1 points with as few as 10 examples. Our contributions are summarized below: 
\setlist{nolistsep}
\begin{enumerate}[leftmargin=*,wide=0pt,noitemsep]
\setlength{\itemsep}{0pt}
\itemsep0em 
    \item 
    We release \sys, the first \pretrained generation model for multi-document inputs with focus on summarization. %
    \item We propose Entity Pyramid, a novel \pretraining strategy 
    that trains the model to select and aggregate salient information from  documents.
    \item 
    We extensively evaluate \sys on 6 datasets from 3 different domains for zero-shot, few-shot and fully-supervised settings. We show that \sys outperforms current state-of-the-art on most of these evaluations with large margins.

\end{enumerate}

\section{Model}

In this section, we discuss our proposed model \sys, a new \pretrained general model for multi-document summarization. Unlike prior work, \sys minimizes dataset-specific modeling by simply concatenating a set of documents and processing them with a general efficient encoder-decoder transformer model (\S\ref{sec:structure}). The underlying transformer model is \pretrained on an unlabeled multi-document dataset, with a new entity-based sentence masking objective to capture the salient information within a set of related documents (\S\ref{sec:objective}).

\subsection{Model Architecture and Input Structure}
\label{sec:structure}
Our goal is to minimize dataset-specific modeling to leverage general \pretrained transformer models for the multi-document task and make it easy to use in practice. Therefore, to summarize a set of related documents, we simply concatenate all the documents in a single long sequence, and process them with an encoder-decoder transformer model. Since the concatenated sequence is long, instead of more standard encoder-decoder transformers like BART~\cite{lewis-etal-2020-bart} and T5~\cite{t5}, we use the Longformer-Encoder-Decoder (LED) Model~\cite{longformer}, an efficient transformer model with linear complexity with respect to the input length.\footnote{ %
We use LED and not other efficient transformers like Bigbird-\pegasus~\cite{bigbird} 
for two reasons, the first is that BigBird's global attention can't be assigned to individual tokens in the middle of the sequence, which is important for the representation of long documents as shown in~\citet{cdlm}. Second, 
because \pretrained checkpoints are available for LED, while BigBird-\pegasus released the already fine-tuned checkpoints.
}
LED uses a sparse local+global attention mechanism in the encoder self-attention side while using the full attention on decoder and cross-attention.

When concatenating, we add special document separator tokens (\texttt{<doc-sep>}) between the documents to make the model aware of the document boundaries (Figure \ref{fig:model_structure}). We also assign 
global attention to these tokens which the model can use to share information across documents~\cite{cdlm}
(see \S\ref{sec:ablation1} for ablations of the effectiveness of this input structure and global attention).

\subsection{Pretraining objective}
\label{sec:objective}

In summarization,
task-inspired \pretraining objectives have been shown to provide gains over general-purpose \pretrained transformers \citep[\pegasus;][]{pegasus}.
In particular, \pegasus introduces Gap Sentence Generation (GSG) as a \pretraining objective where some sentences are masked in the input and the model is tasked to generate them. Following \pegasus, we use the GSG objective, but introduce a new masking strategy designed for multi-document summarization.
As in GSG, we select and mask out $m$ summary-like sentences from 
the input documents
we want to summarize, i.e. every selected sentence is replaced by a single token \texttt{[sent-mask]} in the input, and train the model to generate the concatenation of those sentences as a ``pseudo-summary'' (Figure~\ref{fig:model_structure}). 
This is close to abstractive summarization
because the model needs to reconstruct the masked sentences using the information 
in the rest of the documents.

The key idea 
is how to select sentences that best summarize or represent 
a set of related input documents (which we also call a ``cluster''), not just a single 
document as in standard GSG.
\citet{pegasus} use three strategies - Random, Lead (first $m$ sentences), and ``Principle''. 
The ``Principle'' method computes sentence salience score based on ROUGE score of each sentence, $s_i$, w.r.t the rest of the document ($D/\{s_i\}$), i.e. $\mathrm{Score(}s_i) = \mathrm{\textsc{Rouge}}(s_i,D/\{s_i\})$. 
Intuitively, this assigns a high score to the sentences that have a high overlap with the other sentences.

However, we argue that a naive extension of such strategy to multi-document summarization would be sub-optimal since multi-document inputs typically include redundant information, and such strategy would prefer an exact match between sentences, resulting in a selection of less representative information.

\begin{figure}[tb]
    \centering
    \scriptsize
    \begin{tabular}{@{}p{\linewidth}@{}}
    \toprule
    \textbf{Document \#1}
    Wildfires have burned across tens of thousands of acres of parched terrain in \textcolor{blue}{Colorado}, spurring thousands of evacuations ...\textcolor{blue}{(0.107)}..., residents have sought shelter in middle schools, and local officials fear tourists usually drawn to the region for the summer may not come.\\
    \hline
    \textbf{Document \#2}
    ... \textit{In \textcolor{blue}{Colorado}’s southwest, authorities have shuttered the San Juan National Forest in southwestern Colorado and residents of more than 2,000 homes were forced to evacuate.\textcolor{blue}{(0.187)}}
No homes had been destroyed ... \textit{\textcolor{red}{“Under current conditions, one abandoned campfire or spark could cause a catastrophic wildfire, ..., with human life and property,” said San Juan National Forest Fire Staff Officer Richard Bustamante}}...\\
\hline
\textbf{Document \#3}
The Buffalo Fire west of Denver is ... Several wildfires in \textcolor{blue}{Colorado} have prompted thousands of home evacuations ...\textcolor{blue}{(0.172)}... Nearly 1,400 homes have been evacuated in Summit County, \textcolor{blue}{Colorado}, ...\textcolor{blue}{(0.179)}... \textcolor{red}{“Under current conditions, one abandoned campfire or spark could cause a catastrophic wildfire, ... , with human life and property,” said Richard Bustamante, SJNF forest fire staff officer} ...\\
\hline
\textbf{Entities with High Frequency} \\
\hline
\textcolor{blue}{Colorado}, 416, Tuesday, Wildfires, San Juan National Forest,...\\
\bottomrule
    \end{tabular}
    \caption{An example on sentence selection by Principle vs our Entity Pyramid strategy. Italic text in \textcolor{red}{red} is the sentence with the highest Principle ROUGE scores, which is thereby chosen by the Principle Strategy. Most frequent entity 'Colorado' is shown with \textcolor{blue}{blue}, followed by the Pyramid ROUGE scores in parenthesis. The final selected sentence by Entity Pyramid strategy is in \textit{italic}. which is a better pseudo-summary than the ones selected by the Principle strategy.}
    \label{fig:example}
\end{figure}

 For instance, Figure~\ref{fig:example} shows an example of sentences picked by the Principle strategy \cite{pegasus} vs our Entity Pyramid approach. The figure shows a cluster containing three news articles discussing a wildfire happened in Corolado, and the pseudo-summary of this cluster should be related to the location, time and consequence of the wildfire, but with the Principle strategy, the non-salient sentences quoting the words from an officer are assigned the highest score, as the exact same sentence appeared in two out of the three articles.
In comparison, instead of the quoted words, our strategy selects the most representative sentences in the cluster with high frequency entities.

To address this limitation, we propose a new masking strategy inspired by the Pyramid Evaluation framework~\cite{nenkova-passonneau-2004-evaluating} which was originally developed for evaluating summaries with multiple human written references. Our strategy aims to select sentences that best represent the entire cluster of input documents.

\subsubsection{Entity Pyramid Masking}

\begin{figure*}[tb]
    \centering
    \includegraphics[width=\linewidth]{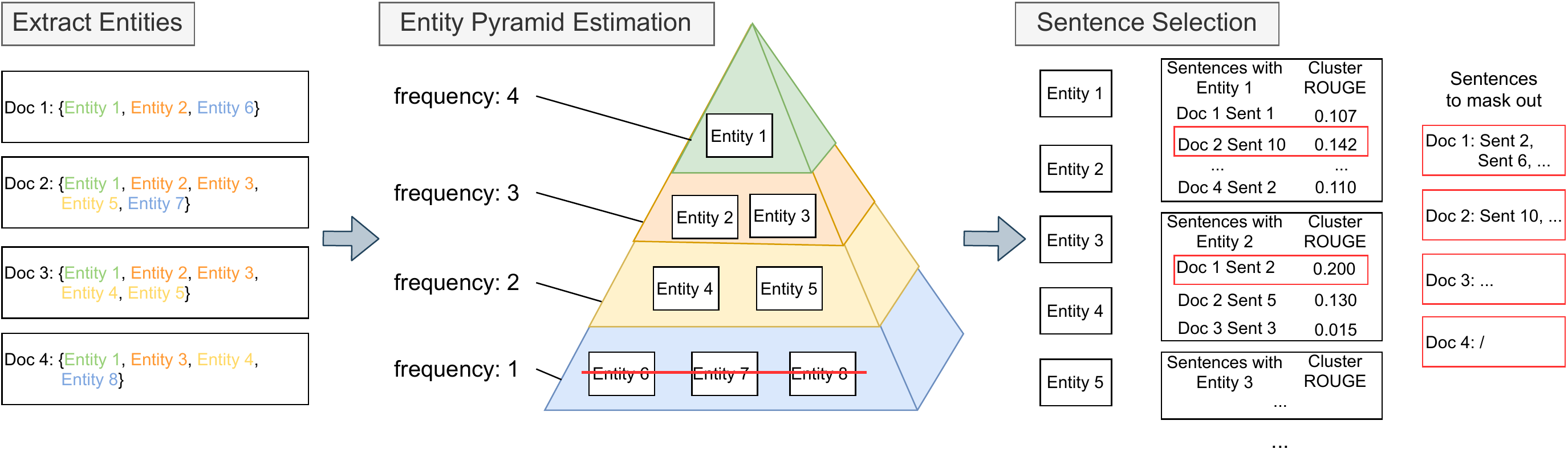}
    \caption{The Entity Pyramid Strategy to select salient sentences for masking.
    Pyramid entity is based on the frequency of entities in the documents.
    The most representative sentence are chosen based on Cluster ROUGE for each entity with frequency $>1$, e.g. Sentence 10 in Document 2 for Entity 1. 
    }
    \label{fig:pyramid}
\end{figure*}

\paragraph{Pyramid Evaluation}
The Pyramid Evaluation method~\cite{nenkova-passonneau-2004-evaluating} is based on the intuition that relevance of a unit of information can be determined by the number of references (i.e. gold standard) summaries that include it. 
The unit of information is called Summary Content Unit (SCU); words or phrases that represent single facts. These SCUs are first identified by human annotators in each reference summary, and they receive a score proportional to the number of reference summaries that contain them. A Pyramid Score for a candidate summary is then the normalized mean of the scores of the SCUs that it contains. One advantage of the Pyramid method is that it directly assesses the content quality.

\paragraph{Entity Pyramid Masking} Inspired by how content saliency is measured
in the Pyramid Evaluation, we hypothesize that a similar idea could be applied in multi-document summarization to identify salient sentences for masking. 
Specifically, for a cluster with multiple related documents, the more documents %
an SCU appears in, the more 
salient that information should be to the cluster. Therefore, it should be considered for inclusion in the pseudo-summary in our masked sentence generation objective. However, SCUs in the original Pyramid Evaluation are human-annotated, which is not feasible %
for large scale \pretraining. 
As a proxy, we explore leveraging information expressed as named entities, since they are key building blocks in extracting information from text about events/objects and the relationships between their participants/parts~\cite{j&m}. Following the Pyramid framework, we use the entity frequency in the cluster as a proxy for saliency.
Concretely, as shown in Fig. \ref{fig:pyramid}, we have the following three steps to select salient sentences in our masking strategy:

\begin{algorithm}[tb]
\footnotesize
\caption{Entity Pyramid Sentence Selection}\label{alg:cap}
\begin{algorithmic}[1]
\Require Document cluster
\Require List of entities w/ frequency $> 1$. $N$ length of the list
\Require $m$ number of sentences to select 
\Ensure List of sentences to mask
\State $E \gets $ sort entities by frequency, descending
\State $selected=[]$
\For{$i \gets 1$ \textbf{to} $|E|$}
    \State $SentCand \gets $ all sentences in the cluster containing $E[i]$
    \State $cur\_sent = \arg\max_{s \in SentCand} Score(s)$
            
    \State $selected.append(cur\_sent)$
    \If{$|selected|==m$}
    \State Break
    \EndIf
\EndFor
\State $\mathrm{\textbf{Return}}$ $selected$
\end{algorithmic}
\end{algorithm}

\begin{enumerate}[leftmargin=*,wide=0pt]
    \setlength\itemsep{0em}
    \item \emph{Entity Extraction}. We extract named entities using SpaCy \cite{spacy}.\footnote{Note that entity information is only used at \pretraining time. This is unlike some prior work (e.g. \citet{pasunuru-etal-2021-efficiently}) that utilize additional information (like named entities, coref, discourse, or AMR) at fine-tuning and inference time.}
    \item \emph{Entity Pyramid Estimation}. We then build an Entity Pyramid for estimating the salience of entities based on their document frequency, i.e. the number of documents each entity appears in.
    \item \emph{Sentence Selection}. Similar to the Pyramid evaluation framework, we identify salient sentences with respect to the cluster of related documents. Algorithm \ref{alg:cap} shows the sentence selection procedure.
    As we aim to select the entities better representing the whole cluster instead of a single document, we first remove all entities from the Pyramid that appear only in one document. Next, we iteratively select entities from top of the pyramid to bottom (i.e., highest to lowest frequency), and then select sentences in the document that include the entity as the initial candidate set. Finally, within this candidate set, we find the most representative sentences to the cluster by measuring the content overlap of the sentence w.r.t documents other than the one it appears in. %
    This final step supports the goal of our \pretraining objective, namely %
    to reconstruct sentences that can be recovered using information from other documents in the cluster, which encourages the model to better connect and aggregate information across multiple documents.
    Following \citet{pegasus} we use ROUGE scores~\cite{rouge} as a proxy for content overlap. For each sentence $s_i$, we specifically define a Cluster ROUGE score as 
    $Score(s_i) = \sum_{\{doc_j\in C, s_i \not\in\ doc_j\}} \mathrm{\textsc{rouge}}(s_i,doc_j)$
    Where $C$ is the cluster of related documents. 

Note that different from the importance heuristic defined in \pegasus~\cite{pegasus}, Entity Pyramid  strategy favors sentences that are representative of more documents in the cluster than the exact matching between fewer documents
(See Figure~\ref{fig:example} for a qualitative example.)
. The benefit of our strategy is shown in an ablation study (\S\ref{sec:ablation1}).

\end{enumerate}

\section{Experiment Goals}
We aim to answer the following questions:
\begin{itemize}[leftmargin=*,wide=0pt]
    \item Q1: How does \sys perform, compared with existing pre-trained generation models in zero- and few-shot settings? See \S\ref{sec:zero_few_shot_evaluation}. 
    \item Q2: How does \sys perform, compared with current state-of-the-art models, in the fully supervised setting? See \S\ref{sec:fully_supervised}.
    \item Q3: How much is the contribution of each component in \sys, i.e. input structure, pretraining, and masking strategy? See \S\ref{sec:ablation1}.
    \item Q4: What is the effect of our entity pyramid strategy, compared with the strategy used in PEGASUS? See \S\ref{sec:ablation1}.
    \item Q5: Is \sys able to capture salient information and generate fluent summaries? See \S\ref{sec:human_eval}.
\end{itemize}
With these goals, we explore the effectiveness of \sys quantitatively on multi-document summarization benchmarks,
verify the improvements by comparing \sys with multiple existing pretrained models and SOTA models, and further validate the contribution of each component with carefully controlled ablations. An additional human evaluation is conducted to show \sys is able to capture salient information and generate more fluent summaries.
\section{Experiments}

\subsection{Experimental Setup}

\paragraph{Implementation Details} %
We use the Longformer-Encoder-Decoder (LED) ~\cite{longformer} large as our model initialization, 
The length limits of input and output are 4096 and 1024, respectively, with sliding window size as $w=512$ for local attention in the input. (More implementation details of pretraining process can be found in Appx \S\ref{sec:detailed_implementation})

\noindent\textbf{Pretraining corpus}
For \pretraining, our goal is to use a large resource where each instance is a set of related documents without any ground-truth summaries. The \emph{Newshead} dataset~\cite{newshead} (row 1, Table \ref{tab:statistics_dataset}) is an ideal choice; it is a relatively large dataset, where every news event is associated with multiple news articles.

\begin{table}[]
    \centering
    \footnotesize
    \setlength{\tabcolsep}{1pt}
    \renewcommand{\arraystretch}{0.85}
    \begin{tabular}{@{}lrrrrr@{}}
    \toprule
     Dataset    & \#Examples&\#Doc/C & $Len_{\mathrm{src}}$ & $Len_{\mathrm{summ}}$\\
     \midrule
     Newshead \citeyearpar{newshead}  &360K&3.5&1734&-\\
     \midrule
    Multi-News \citeyearpar{fabbri-etal-2019-multi}    &56K&2.8&1793&217\\
    Multi-Xscience \citeyearpar{lu-etal-2020-multi-xscience} &40K&4.4&700&105\\
    Wikisum* \citeyearpar{wikisum} &1.5M&40&2238&113\\
    WCEP-10 \citeyearpar{gholipour-ghalandari-etal-2020-large} &10K&9.1&3866&28\\
    DUC2004 \citeyearpar{dang2005overview} &50&10&5882&115\\
    arXiv \citeyearpar{cohan-etal-2018-discourse} &214K&5.5&6021&272\\
    \bottomrule
    \end{tabular}
    \caption{The statistics of all the datasets we explore in this paper. *We use subsets of Wikisum (10/100, 3200) for few-shot training and testing only. 
    }
    \label{tab:statistics_dataset}
\end{table}

\begin{table*}[t]
    \centering
    \scriptsize
    \setlength{\tabcolsep}{4pt}
    \begin{tabular}{@{}lrrrrrrrrrrrrrrrrrr@{}}
    \toprule
    \multirow{2}{*}{Models}     &\multicolumn{3}{c}{Multi-News(256)}&\multicolumn{3}{c}{Multi-XSci(128)}& \multicolumn{3}{c}{WCEP(50)} &\multicolumn{3}{c}{WikiSum(128)} & \multicolumn{3}{c}{arXiv(300)}& \multicolumn{3}{c}{DUC2004 (128)}\\
    \cmidrule(lr){2-4} \cmidrule(lr){5-7} \cmidrule(lr){8-10}\cmidrule(lr){11-13} \cmidrule(lr){14-16}\cmidrule(lr){17-19}
                 & R-1 & R-2 & R-L & R-1 & R-2 & R-L& R-1 & R-2 & R-L & R-1 & R-2 & R-L& R-1 & R-2 & R-L & R-1 & R-2 & R-L \\
    \midrule
        \textsc{Pegasus}$\star$\cite{pegasus} & 36.5 &10.5 &18.7 &-&-&-&-&-&-&-&-&-& 28.1 & 6.6 &17.7 &-&- &-\\    

               \pegasus (our run)  & 32.0 & 10.1 & 16.7 & 27.6 & \textbf{4.6} & 15.3 &\textbf{33.2}&\textbf{12.7}&\textbf{23.8} &24.6&5.5&15.0& 29.5& 7.9 &17.1 & 32.7 &\textbf{7.4} & 17.6 \\
        BART (our run) &27.3 &6.2&15.1 &18.9& 2.6& 12.3 &20.2 &5.7 &15.3 &21.6 &5.5 &15.0 &29.2 &7.5 &16.9 &24.1 &4.0 &15.3 \\
        \hhline{===================}
        LED (our run) &17.3 &3.7 &10.4 &14.6 &1.9 &9.9 &18.8 &5.4 &14.7 &10.5 &2.4 &8.6 &15.0 &3.1 &10.8 &16.6 &3.0 &12.0\\
        \sys (our model)&\textbf{42.0}&\textbf{13.6}&\textbf{20.8}&\textbf{29.1}&4.6&\textbf{15.7}&28.0 &10.3 &20.9 &\textbf{28.0} &\textbf{8.0} &\textbf{18.0} &\textbf{34.6} &\textbf{9.4} &\textbf{18.3} &\textbf{35.1} &7.2&\textbf{17.9}\\    
    \bottomrule
    \end{tabular}
    \caption{Zero-shot results. The models in the first block use the full-length attention ($O(n^2)$) and are \pretrained on the single document datasets. The numbers in the parenthesis following each dataset indicate the output length limit set for inference. \textsc{Pegasus}$\star$ means results taken exactly from \pegasus \cite{pegasus}, where available.}
    \label{tab:zero-shot}
\end{table*}

\noindent\textbf{Evaluation Datasets} We evaluate our approach on wide variety of multi-document summarization datasets plus one single document dataset from various domains (News, Wikipedia, and Scientific literature). See Table~\ref{tab:statistics_dataset} for dataset statistics and Appx. \S\ref{sec:datasets} for details of each dataset.

\noindent\textbf{Evaluation metrics} Following previous works~\cite{pegasus}, we use ROUGE scores (R-1, -2, and -L), which are the standard evaluation metrics, to evaluate the downstream task of multi-document summarization.\footnote{We use https://github.com/google-research/google-research/tree/master/rouge with default stemmer settings.} For better readability, we use AVG ROUGE scores (R-1, -2, and -L) for evaluation in the few-shot setting.
\subsection{Zero- and Few-shot Evaluation}
\label{sec:zero_few_shot_evaluation}
Many existing works in adapting \pretrained models for summarization require large amounts of fine-tuning data, which is often impractical for new domains. In contrast, since our \pretraining strategy is mainly designed for multi-document summarization, we expect that our approach can quickly adapt to new datasets without the need for significant fine-tuning data. To test this hypothesis, we first provide evaluation results in zero and few-shot settings where the model is provided with no, or only a few (10 and 100) training examples. Obtaining such a small number of examples should be viable in practice for new datasets.

\paragraph{Comparison}
To better show the utility of our \pretrained models, we compare with three state-of-the-art \pretrained generation models: 
BART~\cite{lewis-etal-2020-bart}\footnote{
Pilot experiments comparing BART and T5 showed BART to outperform T5 on the few-shot evaluation
of Multi-News (with AVG ROUGE of 23.5/26.4 (T5) v.s. 25.2/26.7 (BART) for 10/100 training examples, respectively). Thus, we are using BART as one of the baselines.},
PEGASUS~\cite{pegasus} and Longformer-Encoder-Decoder(LED)~\cite{longformer}. These \pretrained models have been shown to outperform dataset-specific models in summarization~\cite{lewis-etal-2020-bart,pegasus}, and because of \pretraining, they are expected to also work well in the few-shot settings.
As there is no prior work doing few-shot and zero-shot evaluations on all the datasets we consider, and also the results in the few-shot setting might be influenced by sampling variability (especially with only 10 examples)~\cite{bragg2021flex}, we run the same experiments for the compared models five times with different random seeds (shared with all the models), 
 with the publicly available checkpoints
.\footnote{Checkpoints from  https://huggingface.co/models}

\begin{figure*}[t]
    \centering
    \includegraphics[width=\linewidth]{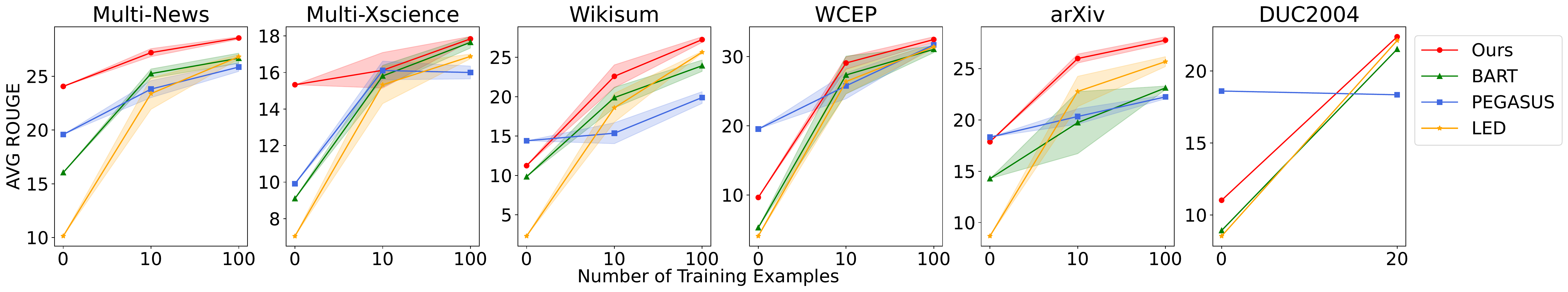}
    \caption{
    The AVG ROUGE scores (R-1, R-2 and R-L) of the \pretrained models with 0, 10 and 100 training data with variance. All the results of few-shot experiments (10 and 100) are obtained by the average of 5 random runs (with std, and the same set of seeds shared by all the models). Note that DUC2004 only has 50 examples, we use 20/10/20 for train/valid/test in the few-shot experiments.}
    \label{fig:fewshot}
\end{figure*}

Similar to \citet{pasunuru-etal-2021-efficiently}, the inputs of all the models are the concatenations of the documents within the clusters (in the same order), each document is truncated based on the input length limit divided by the total number of documents so that all documents are represented in the input. \footnote{Pilot experiments show simple truncation results in inferior performance, which is in line with \citet{pasunuru-etal-2021-efficiently}.} 

To preserve the same format as the corresponding \pretrained models, we set the length limit of output
for BART and \pegasus exactly as their \pretrained settings on all of the datasets (except for the zero-shot experiments, the details can be found in Sec.\ref{sec:zeroshot}). Regarding length limit of inputs, we tune the baselines by experimenting with 512, 1024, 4096 on Multi-News dataset in few-shot setting (10 data examples), and the model with length limit 512(\pegasus)/1024(BART) achieves the best performance, thus we use this setting (detailed experiment results for different input lengths can be found in Appx. \S\ref{sec:input_length_limit}).
We use the same length limit as our model for the LED model, i.e. 4096/1024 for input and output respectively, for all the datasets. 

\subsection{Zero-Shot Results}
\label{sec:zeroshot}

For zero-shot\footnote{For clarity, by zero-shot we mean using the pretrained model directly without any additional supervision.} abstractive summarization experiments, since the models have not been trained on the downstream datasets, the lengths of generated summaries mostly depend on the \pretrained settings. Thus to better control the length of generated summaries and for a fair comparison between all models, following \citet{Zhu2019MakeLB}, we set the length limit of the output at inference time to the average length of gold summaries.\footnote{In practice, it is reasonable to assume knowing the approximate length of the expected summary for a given task/domain.} Exploring other approaches to controlling length at inference time \citep[e.g.,][]{wu2021controllable} is an orthogonal direction, which we leave for future work.

Table \ref{tab:zero-shot} shows the performance comparison among all the models. Results indicate that our model achieves substantial improvements compared with all the three baselines on most of the datasets. As our model is \pretrained on clusters of documents with longer input and output, the benefit is stronger on the dataset with longer summaries, e.g. Multi-News and arXiv. Comparing \pegasus and BART models, as the objective of \pegasus is designed mainly for summarization tasks, not surprisingly it has relatively better performances across different datasets. 
Interestingly, LED underperforms other models, plausibly since part of the positional embeddings (1k to 4k) are not \pretrained. Encouragingly, our model performs the best, demonstrating the benefits of our \pretraining strategy for multi-document summarization.
\subsection{Few Shot Evaluation}

Compared with the strict zero-shot scenario, few-shot experiments are closer to the practical scenarios, as it is arguably affordable to label dozens of examples for almost any application. 

We fine-tune all of the four models on different subsets with 10 and 100 examples, and the results are shown in Figure~\ref{fig:fewshot}. (hyperparameter settings in Appx. \S\ref{sec:fewshot_setting}) Since R-1, -2, and -L show the same trend, we only present the average of the three metrics in the figure for brevity (full ROUGE scores can be found in Appx. Table~\ref{tab:rouge_score_fewshot})
To show the generality, all the results of few-shot experiments are the average over 5 runs on different subsets (shared by all the models).

The result of each run is obtained by the `best' model chosen based on the ROUGE scores on a randomly sampled few-shot validation set with the same number of examples as the training set, which is similar with \citet{pegasus}. 
Note that their reported best models have been selected based on the whole validation set which may give \pegasus some advantage. Nevertheless, we argue that sampling few-shot validation sets as we do here is closer to real few-shot scenarios \cite{bragg2021flex}.

Our model outperforms all baselines on all of the datasets with 10 and 100 examples demonstrating the benefits of our \pretraining strategy and input structure. Comparing the performances of our model with the different number of training data fed in, our model converges faster than other models with as few as 10 data examples.

\subsection{Fully Supervised Evaluation}
\label{sec:fully_supervised}

\footnotetext{We re-evaluate the generated summaries of the models from \citet{lu-etal-2020-multi-xscience} for Multi-XScience, as we use a different version of ROUGE.}
To show the advantage of our \pretrained model when there is abundant training data, we also train the model with the full training set (hyperparameter settings can be found in Appx. \S\ref{sec:fullsupervised_setting}). Table~\ref{tab:full_supervisory_all} shows the performance comparison with previous state-of-the-art\footnote{Due to the lack of computational resources, we do not train the model on Wikisum.}, along with the results of previous SOTA. 
\begin{table}[t]
    \centering
    \footnotesize
    \setlength{\tabcolsep}{4pt}
    \begin{tabular}{@{}lrrrrrr@{}}
    \toprule
    \multirow{2}{*}{Datasets}     & \multicolumn{3}{c}{Previous SOTA} & \multicolumn{3}{c}{\sys} \\
    \cmidrule(lr){2-4} \cmidrule(lr){5-7}
                 & R-1 & R-2 & R-L & R-1 & R-2 & R-L \\
    \midrule
    Multi-News     & 49.2 &19.6&24.5  &\textbf{49.9}&\textbf{21.1}&\textbf{25.9}\\
    Multi-XScience &\textbf{33.9}&6.8& \textbf{18.2} &31.9&\textbf{7.4}& 18.0 \\
    WCEP&35.4&15.1&25.6 &\textbf{46.1}&\textbf{25.2}&\textbf{37.9}\\
    arXiv & 46.6&19.6&41.8&
    \textbf{47.6}&\textbf{20.8}&\textbf{42.6}\\    
    \bottomrule
    \end{tabular}
    \caption{Fully supervised results. Previous SOTA are from \citet{pasunuru-etal-2021-efficiently} for Multi-News, \citet{lu-etal-2020-multi-xscience} for Multi-XScience\footnotemark,
    \citet{dyne} for WCEP, and \citet{longformer} for arXiv. 
    }
    \label{tab:full_supervisory_all}
\end{table}
We observe that \sys achieves state-of-the-art results on Multi-News, WCEP, and arXiv, while slightly underperforming the prior work on Multi-XScience (R-1). One possible explanation is that in Multi-XScience clusters have less overlapping information than in the corpus on which \sys was pretrained. In particular, the source documents in this dataset are the abstracts of all the publications cited in the related work paragraphs, which might be less similar to each other and the target related work(i.e., their summary) . \sys outperforms the LED model (State-of-the-art) on the arXiv dataset while using a sequence length 4x shorter (4K in \sys v.s. 16K in LED), further showing that the \pretraining and input structure of our model not only works for multi-document summarization, but can be also effective for summarizing single documents having multiple sections.

\section{Ablation Study}
\label{sec:ablation1}

\begin{figure}
    \centering
    \includegraphics[width=\linewidth]{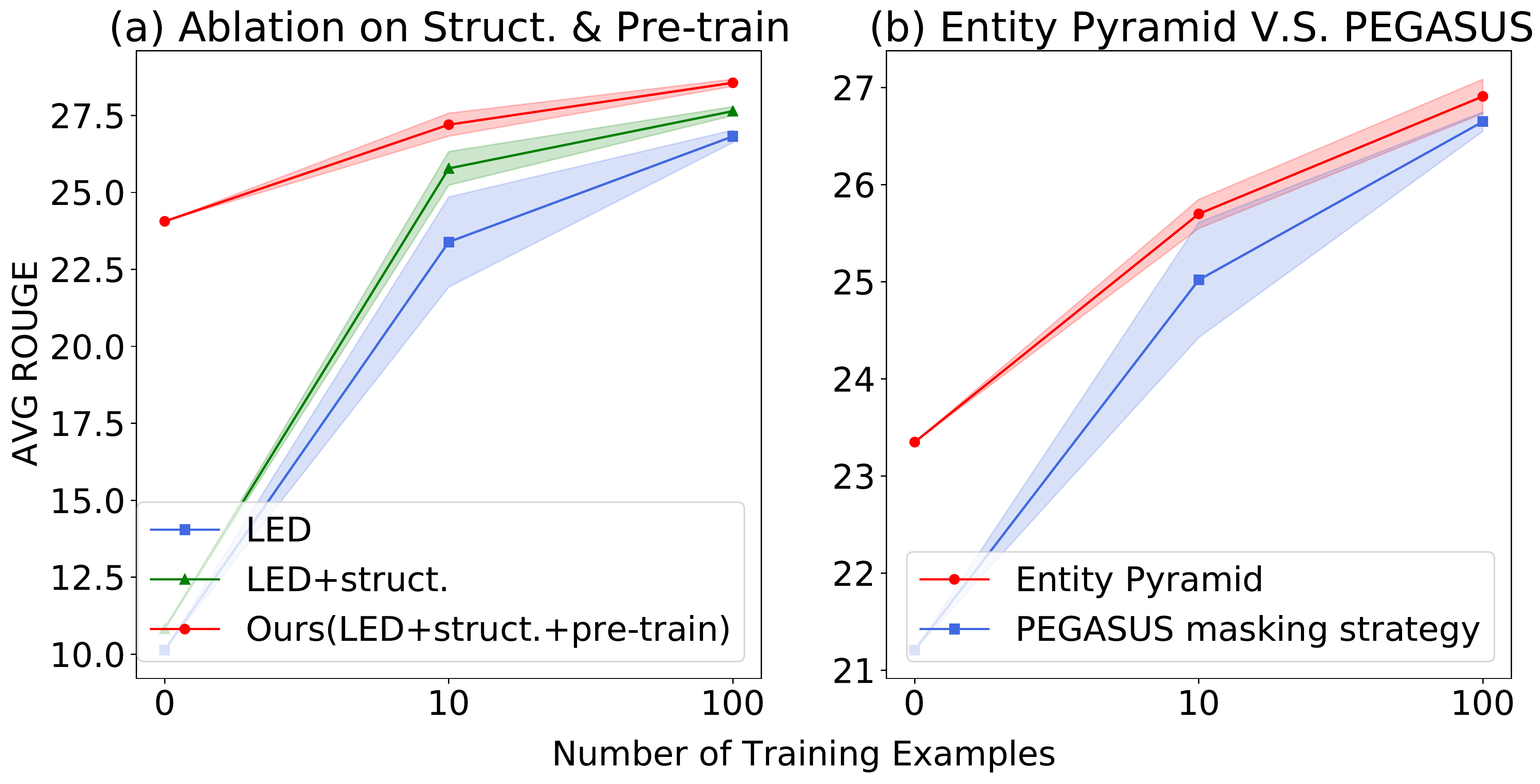}
    \caption{Ablation study with the few-shot setting on the Multi-News dataset regarding to (a) input Structure (\texttt{<doc-sep>} tokens between documents and global attention on them) and \pretraining, (b) \pretraining using \pegasus vs our approach. \vspace{-4mm}
    }
    \label{fig:ablation}
    \vspace{-2mm}
\end{figure}

 We conduct ablation studies on the Multi-News dataset in few-shot setting, to validate the contribution of each component in our \pretrained models. 
 
\noindent \textbf{Input structure:} In Figure \ref{fig:ablation} (a) we observe the effectiveness of both \pretraining and the input structure (\texttt{<doc-sep>} tokens between documents and global attention on them).

\noindent \textbf{Sentence masking strategy:} To isolate the effect of our proposed \pretraining approach, 
we compare with  a model with exactly the same architecture when \pretrained on the same amount of data but using the \pegasus \cite{pegasus} masking strategy instead of ours.
In other words, we keep all the other settings the same (e.g., data, length limit of input and output, \pretraining dataset, input structure, as well as the separators) and only modify the \pretraining masking strategy.
We run the same experiments under zero-/few-shot scenarios on the Multi-News dataset as in \S\ref{sec:zero_few_shot_evaluation}, and the results are shown in 
Figure~\ref{fig:ablation} (b).
The model \pretrained with our Entity Pyramid strategy shows a clear improvement under few-shot scenarios.

\section{Human Evaluation}
\label{sec:human_eval}
We also conduct human evaluations to validate the effectiveness of \sys on DUC2007 and TAC2008~\cite{tac2008} datasets in the few-shot setting (10/10/20 examples for train/valid/test). Both datasets consist of clusters of news articles, and DUC2007 contains longer inputs (25 v.s. 10 documents/cluster) and summaries (250 v.s. 100 words). Since the goal of our method is to enable the model to better aggregate information across documents, we evaluate the content quality of the generated summaries following the original Pyramid human evaluation framework \cite{nenkova-passonneau-2004-evaluating}. In addition, we also evaluate the fluency of generated summaries following the DUC guidelines.\footnote{\url{https://www-nlpir.nist.gov/projects/duc/duc2007/quality-questions.txt}}

\paragraph{Settings}
Three annotators\footnote{We recruited expert annotators with payment above average of the participants’ demographics.} are hired to do both Pyramid Evaluation and Fluency evaluation, they harmonize the standards on one of the examples. Specifically, for each data example, we provide three anonymized system generated summaries, along with a list of SCUs. The annotators are asked to find all the covered SCUs for each summary, and score the fluency in terms of Grammaticality, Referential clarity and Structure \& Coherence, according to DUC human evaluation guidelines, with a scale 1-5 (worst to best). They are also suggested to make comparison between three generated summaries into consideration when scoring the fluency. To control for the ordering effect  of the given summaries, we re-order the three summaries for each data example, and ensure the chance of their appearance in different order is the same (e.g. BART appears as summary A for 7 times, B for 7 times and C for 6 times for both datasets). The instruction for human annotation can be found in Figure~\ref{fig:humaneval_1} and Figure~\ref{fig:humaneval_2} in the appendix. Annotators were aware that annotations will be used solely for computing aggregate human evaluation metrics and reporting in the scientific paper.

\paragraph{Compared Models}
We compare our model with LED and PEGASUS in human evaluations. Because PEGASUS is a task-specific  model for abstractive summarization, and LED has the same architecture and length limits as our model with the parameters inherited from BART, which is more comparable with our model than vanilla BART. 

\begin{table}[]
    \centering
    \footnotesize
    \setlength{\tabcolsep}{4pt}
    \begin{tabular}{@{}lrrrrrrrr@{}}
    \toprule
    \multirow{2}{*}{Model}&\multicolumn{4}{c}{DUC2007(20)}&\multicolumn{4}{c}{TAC2008(20)}\\
& $S_r$& R & P & F& $S_r$& R & P & F\\
    \midrule
    PEGASUS&6.0&2.5&2.4&2.4&\textbf{8.7}&\textbf{9.1}&9.4&9.1\\
    LED&9.6&3.9&4.0&3.8&6.9&7.1&\textbf{10.8}&8.4\\
    \sys&\textbf{12.5}&\textbf{5.1}&\textbf{5.0}&\textbf{5.0}&8.5&8.9&10.0&\textbf{9.3}\\
    \bottomrule
    \end{tabular}
    \caption{Pyramid Evaluation results: Raw scores $S_r$, (R)ecall, (P)recision and (F)-1 score. For readability, Recall, Precision and F-1 scores are multiplied by 100.}
    \label{tab:pyramid_evaluation}
\end{table}

\paragraph{Pyramid Evaluation}
Both TAC and DUC datasets include SCU (Summary Content Unit) annotations and weights identified by experienced annotators. We then ask 3 annotators to make a binary decision whether each SCU is covered in a candidate summary. Following \citet{nenkova-passonneau-2004-evaluating}, the raw score of each summary is then computed by the sum of weights of the covered SCUs, i.e. $S_r = \sum_{SCU} w_iI(SCU_i)$, where $I(SCU_i)$ is an indicator function on whether $SCU_i$ is covered by the current summary, and $w_i$ is the weight of $SCU_i$. 
In the original pyramid evaluation, the final score is computed by the ratio of $S_r$ to the maximum possible weights with the same number of SCUs as in the generated summaries. However,  the total number of SCUs of generated summaries is not available in the simplified annotations in our design. To take consideration of the length of generated summaries and make a fair comparison, instead, we compute Recall, Precision and F-1 score regarding lengths of both gold references and system generated summaries as
\begin{equation*}\label{eq:pyramid}
\small
\mathrm{R}{=}\frac{S_r}{len(gold)};\;\;
\mathrm{P}{=}\frac{S_r}{len(sys)};\;\;
\mathrm{F}1{=}\frac{2\cdot R \cdot P}{(R+P)}
\end{equation*}

\begin{table}[]
    \centering
    \footnotesize
    \setlength{\tabcolsep}{2.5pt}
    \begin{tabular}{@{}lcccccc@{}}
    \toprule
    \multirow{2}{*}{Model}&\multicolumn{3}{c}{DUC2007(20)}&\multicolumn{3}{c}{TAC2008(20)}\\
& Gram.&Ref.&		Str.\&Coh.		&Gram.&Ref.&		Str.\&Coh.\\
    \midrule
    PEGASUS&4.45&4.35&1.95&\textbf{4.40}&4.20&3.20\\
    LED&4.35&4.50&3.20&3.10&3.80&2.55\\
    \sys&\textbf{4.70}&\textbf{4.65}&\textbf{3.70}&\textbf{4.40}&\textbf{4.45}&\textbf{4.10}\\
    \bottomrule
    \end{tabular}
    \caption{The results of Fluency Evaluation on two datasets, in terms of the Grammaticality		, Referential clarity and Structure \& Coherence.\vspace{-6pt}}
    \label{tab:fluency_eval}
\end{table}
\paragraph{Fluency Evaluation}
Fluency results can be found in Table~\ref{tab:fluency_eval}, 
and \sys has the best performance on both datasets in terms of all aspects. Only for Grammaticality  PRIMERA's top performance is matched by PEGASUS.

\section{Related Work}
\label{sec:related}

\paragraph{Neural Multi-Document Summarization}
These models can be categorized into two classes, graph-based models~\cite{yasunaga-etal-2017-graph,liao-etal-2018-abstract,li-etal-2020-leveraging-graph,pasunuru-etal-2021-efficiently} and hierarchical models~\cite{liu-lapata-2019-hierarchical,fabbri-etal-2019-multi,jin-etal-2020-multi}. 
Graph-based models often require auxiliary information (e.g., AMR, discourse structure) to build an input graph, making them reliant on auxiliary models and less general.
Hierarchical models are another class of models for multi-document summarization, examples of which include multi-head pooling and inter-paragraph attention ~\cite{liu-lapata-2019-hierarchical}, MMR-based attention~\cite{fabbri-etal-2019-multi,mao2020multi}, and attention across representations of different granularity (words, sentences, and documents)~\cite{jin-etal-2020-multi}. 
Prior work has also shown the advantages of customized optimization in multi-document summarization \citep[e.g., RL;][]{su2021pobrl}.
Such models are often dataset-specific and difficult to develop and adapt to other datasets or tasks.

\paragraph{Pretrained Models for Summarization}
Pretrained language models have been successfully applied to summarization, e.g., BERTSUM~\cite{liu-lapata-2019-text}, BART~\cite{lewis-etal-2020-bart}, T5~\cite{t5}.
Instead of regular language modeling objectives, \pegasus~\cite{pegasus} introduced a pretraining objective with a focus on summarization, using Gap Sentence Generation, where the model is tasked to generate summary-worthy sentences, and \citet{zou-etal-2020-pre}  proposed different pretraining objectives to reinstate the original document, specifically for summarization task as well. Contemporaneous work by \citet{rothe-etal-2021-thorough}
argued that task-specific pretraining does not always help for summarization, however, their 
experiments are limited to single-document summarization datasets.
Pretraining on the titles of HTMLs has been recently shown to be useful for few-shot short-length single-document summarization as well \cite{HTLM}.
\citet{goodwin-etal-2020-flight}  evaluate three state-of-the-art models (BART, \pegasus, T5) on several multi-document summarization datasets with low-resource settings, showing that 
abstractive multi-document summarization remains challenging. 
Efficient pretrained transformers (e.g., Longformer \cite{longformer} and BigBird \cite{bigbird} that can process long sequences have been also proven successful in summarization, typically by the ability to process long inputs, connecting information across the entire sequence. 
CDLM \cite{cdlm} is a follow-up work for pretraining the Longformer model in a cross-document setting using global attention on masked tokens during pretraining. However, this model only addresses encoder-specific tasks and it is not suitable for generation. 
In this work, we show how efficient transformers can be pretrained using a task-inspired pretraining objective for multi-document summarization. 
Our proposed method is also related to the PMI-based token masking \citet{levine2020pmi} which improves over random token masking outside summarization.

\section{Conclusion and Future Work}
In this paper, 
we present \sys a pre-trained model for multi-document summarization.
Unlike prior work, \sys minimizes dataset-specific modeling by using a Longformer model \pretrained with a novel entity-based sentence masking objective. The \pretraining objective is designed to help the model connect and aggregate information across input documents.
\sys outperforms prior state-of-the-art pre-trained and dataset-specific models on 6 summarization 
datasets from 3 different domains, on zero, few-shot, and full fine-tuning setting. 
\sys's top performance is also revealed by human evaluation.

In zero-shot setting, we can only control the output length of generated summaries at inference time by specifying a length limit during decoding. Exploring a controllable generator in which the desired length can be injected as part of the input is a natural future direction. Besides the summarization task, we would like to explore using \sys for other generation tasks with multiple documents as input, like multi-hop question answering.
\section*{Ethics Concern}
While there is limited risk associated with our work, similar to existing state-of-the-art generation models, there is no guarantee that our model will always generate factual content. Therefore, caution must be exercised when the model is deployed in practical settings. Factuality is an open problem in existing generation models.

\bibliography{custom}
\bibliographystyle{acl_natbib}
\appendix
\section{Implementation details of pre-training}
\label{sec:detailed_implementation}
As the multi-document summarization task has a higher compression ratio, defined as $len(Summary)/len(Input)$, 
(e.g. 12\% for Multi-News dataset and 15\% for Multi-Xscience dataset), we use 15\% as the ratio of masked sentences for generation. In addition to this 15\% masked sentences, following \pegasus~\cite{pegasus}, we also copy an additional 15\% of the input sentences to the output without masking them in the input. This allows the model to also learn to copy information from the source directly and found to be useful by \citet{pegasus}.

We \pretrain the model for 100K steps, with early stopping, batch size of 16, Adam optimizer with a learning rate of $3e{-}5$ following \citet{longformer}, with 10K warmup steps and linear decay. The pretraining process takes likely 7 days on 4 A100 GPUs.

As the backbone of \sys is the Longformer Encoder Decoder model (LED), it has the same number of parameters with LED (447M). 
\section{Detailed Description on the Evaluation Datasets}
\label{sec:datasets}
The details of evaluation datasets can be found below.

\emph{Multi-News} \cite{fabbri-etal-2019-multi}: A multi-document dataset with summaries written by professional editors from the newser.com.

\emph{Wikisum} \cite{wikisum} Each summary is a Wikipedia article, and the source documents are either citations in the reference section or the Web Search results of section titles.\footnote{Due to the large size of the dataset, we evaluate all the models on the first 3200 data in the test set. And in the few-shot experiments, we randomly choose few examples (10 or 100) from the training set and validation set.} In our experiments, we use the data crawled by \citet{liu-lapata-2019-hierarchical}.

\emph{WCEP} \cite{gholipour-ghalandari-etal-2020-large} is built based on news events from Wikipedia Current Events Portal and the references are obtained similar to Wikisum. 
There are at most 100 documents within each cluster in the original dataset, thus we remove all the duplicates and only keep up to 10 documents for each cluster based on the relevance score in the original dataset, which is similar to the WCEP-10 variant in the original paper. 

\emph{Multi-X-Science} \cite{lu-etal-2020-multi-xscience} a multi-document summarization dataset created from scientific articles, the summaries are paragraphs of related work section, while source documents include the abstracts of the query and referred papers. 

\emph{DUC} benchmarks \cite{dang2005overview} include multi-document summarization datasets in the news domain, with 10-30 documents and 3-4 human-written summaries per cluster. Since these datasets are small, we use them primarily for a few-shot evaluation. We use DUC2003 for training (only one of the reference summaries for each document is used for training) and DUC2004 as test. 

\emph{ArXiv} \cite{cohan-etal-2018-discourse} is a single document summarization dataset in the scientific paper domain. Each document is a scientific paper, and the summary is the corresponding abstract. As each scientific paper consists of multiple sections, we treat each section as a separate document within a cluster in our experiments. This is to evaluate our model's effectiveness on summarizing single documents having multiple sections.

\section{Details on Compared models}
The details of compared models in the zero-/few-shot setting can be found below.

\emph{BART} \cite{lewis-etal-2020-bart} an encoder-decoder transformer model \pretrained on the objective of reconstructing the corrupted documents in multiple ways, e.g. Token Deletion, Text Infilling, Sentence Rotation and etc. 

\emph{\pegasus} \cite{pegasus} a \pretrained model designed for abstractive summarization as the downstream task, especially for the single document input. It is trained on the objective of Gap Sentence Generation on C4~\cite{t5} and  Hugenews datasets (Note that the \pretraining data size in \pegasus is magnitudes larger than ours). As it is only evaluated on one multi-document summarization dataset (Multi-news), we rerun the model on all the datasets. To verify the quality of our reproduction, the average ROUGE scores of our re-run model vs. (the ones reported on the paper) with 10 examples and 100 examples fed are $23.81\pm0.79$ vs. (24.13) and $25.86\pm0.41$ vs. (25.48), with minor differences plausibly resulting from different samplings.

\emph{Longformer Encoder-Decoder (LED)} \cite{longformer}
is the initial state of our model before \pretraining. The parameters of LED are inherited from the BART model, and to enable the model to deal with longer input, the position embeddings are repeatedly copied from BART’s 1K position embeddings. It is different from our model with respect to both \pretraining and input structure (document separators and global attentions), with global attention on the (\texttt{<s>}) token only and no document separators.
\subsection{Detailed Experiment for Input Length Limit}
\label{sec:input_length_limit}
We run an experiment to select the proper length limit for compared pretrained models, i.e. BART and \pegasus. Specifically, we train both models with different input length limits (512/1024/4096) in the few-shot setting (with 10 data examples) on the multi-news dataset. Similar as the few-shot experiments described in \S\ref{sec:zero_few_shot_evaluation}, we train each model with each specific input length limit for 5 times on different subsets, which are shared by all the models. As shown in Table~\ref{tab:input_length_experiment}, BART with length limit 1024 performs the best and \pegasus with length limit 512 performs the best, thus in all our experiments, we use 1024 as the input length limit for BART and 512 for \pegasus. 
\begin{table}[]
    \centering
    \setlength{\tabcolsep}{4pt}
    \begin{tabular}{@{}lrrrrrr@{}}
    \toprule
    \multirow{2}{*}{Length Limit}     & \multicolumn{3}{c}{BART} & \multicolumn{3}{c}{\pegasus}\\
    \cmidrule(lr){2-4} \cmidrule(lr){5-7}& R-1 & R-2 & R-L & R-1 & R-2 & R-L \\
    \midrule
    512 & - &-&-&\textbf{39.0}&\textbf{12.1}&\textbf{20.3}\\
    1024& \textbf{42.3}&\textbf{13.7}&\textbf{19.7}&37.6&10.7&18.8\\
    4096& 37.9&11.0&17.5&34.9&8.7&17.6\\
    \bottomrule
    \end{tabular}
    \caption{The ROUGE score (R-1/R-2/R-3) for pretrained models (BART and \pegasus) with different input length limit in few-shot setting (10 data example) on the multi-news dataset. The results are the average over 5 runs on different subsets (the same seeds shared with all the other models in this paper).}
    \label{tab:input_length_experiment}
\end{table}
\section{Hyperparameters in Few-shot and Full Supervised Experiments}
\subsection{Few-shot Experiments}
\label{sec:fewshot_setting}
We use Adam as the optimizer with linear scheduled learning rate $3e-5$ for BART, LED and our model, and use the default optimization settings of the few-shot experiments from \citet{pegasus}, i.e. AdaFactor optimizer with scheduled learning rate $5e-4$. For all the experiments with 10 examples, the batch size is 10, the models are trained for 200 steps, with warm-up as 20 steps. For the experiments with 100 examples, we use the same batch size, with the total step and warm-up step set to be 1000 and 100, respectively. 
\subsection{Fully Supervised Experiments}
\label{sec:fullsupervised_setting}
We use Adam as the optimizer with linear scheduled learning rate $3e-5$, and batch size as 16 for all the datasets in the full supervised experiments. The number of steps and warm-up steps are set based on the size of the datasets. The details can be found in Table~\ref{tab:setting_full_supervised}

\begin{table}[th!]
    \centering
    \resizebox{\linewidth}{!}{
    \begin{tabular}{lrr}
    \toprule
    Dataset & Total Steps & Warmup Steps\\
    \midrule
    Multi-News     &25k  &2.5k\\
     Multi-XScience    &20k &2k\\
     WCEP &5k &.5k \\
     arXiv &40k &4k \\
     \bottomrule
    \end{tabular}}
    \caption{Details of total steps and warm-up steps used in the Full Supervised experiments.}
    \label{tab:setting_full_supervised}
\end{table}
\section{Detailed Results in Few-shot Setting}
\label{sec:detailed_fewshot}
The exact ROUGE scores in Figure~\ref{fig:fewshot} are shown in Table~\ref{tab:rouge_score_fewshot}.
\begin{table}[th!]
    \centering
    \resizebox{\linewidth}{!}{
    \begin{tabular}{l|c|c|c}
    \toprule
    Model     &  0 Examples & 10 Examples & 100 Examples\\
    \hhline{====}
    \multicolumn{4}{c}{Multi-News}\\
    \hline
    \pegasus&31.97/10.06/16.74&39.02/12.10/20.32&42.99/13.50/21.10\\
    BART&26.10/8.98/13.06&42.30/13.74/19.71&44.23/14.77/21.02\\
    LED&16.60/4.78/9.05&38.86/12.48/18.82&44.45/14.85/21.16\\
    Ours     & \textbf{39.09}/\textbf{13.91}/\textbf{19.19}&\textbf{44.02}/\textbf{15.54}/\textbf{22.03}&\textbf{46.01}/\textbf{16.76}/\textbf{22.91}\\
    \hhline{====}
    \multicolumn{4}{c}{Multi-Science}\\
    \hline
    \pegasus&\textbf{27.33}/4.77/\textbf{15.04}&28.14/4.68/\textbf{15.49}&28.01/4.09/15.89\\
BART&15.21/3.49/8.61&27.80/4.74/14.90&31.17/5.32/16.45\\
    LED&11.79/2.47/6.86&26.57/4.05/15.36&29.46/4.85/16.32\\
Ours&26.90/\textbf{4.98}/14.09&\textbf{28.36}/\textbf{4.73}/15.29&\textbf{31.25}/\textbf{5.43}/\textbf{16.84}\\
\hhline{====}
\multicolumn{4}{c}{Wikisum}\\
    \hline
    \pegasus&\textbf{23.67}/\textbf{5.37}/\textbf{14.17}&23.44/6.44/16.21&28.50/9.83/21.33\\
BART&15.80/4.60/9.13&28.95/9.88/20.80&32.97/13.81/25.01\\
    LED&8.70/2.34/5.78&26.53/9.30/19.95&34.15/16.03/26.75\\
Ours&17.79/5.02/10.90&\textbf{31.10}/\textbf{13.26}/\textbf{23.39}&\textbf{36.05}/\textbf{17.85}/\textbf{27.81}\\
\hhline{====}
\multicolumn{4}{c}{WCEP}\\
    \hline
    \pegasus&\textbf{27.69}/\textbf{10.85}/\textbf{20.03}&35.60/14.84/26.84&42.09/19.93/33.04\\
BART&7.11/3.41/5.32&37.46/15.82/28.70&41.34/19.19/32.58\\
LED&5.69/2.19/4.32&36.29/15.04/27.80&41.83/19.46/32.92\\
Ours&13.50/5.30/10.11&\textbf{38.97}/\textbf{17.55}/\textbf{30.64}&\textbf{42.96}/\textbf{20.53}/\textbf{33.87}\\
\hhline{====}
\multicolumn{4}{c}{arXiv}\\
    \hline
\pegasus&\textbf{29.76}/7.94/\textbf{17.27}&33.10/8.52/19.40&36.38/9.55/20.83\\
BART&23.26/7.57/12.01&32.53/8.70/17.98&37.62/10.78/20.99\\
LED&13.94/3.76/8.35&36.51/11.16/20.68&41.00/13.74/22.34\\
Ours&29.14/\textbf{8.64}/15.82&\textbf{41.13}/\textbf{13.81}/\textbf{23.02}&\textbf{43.42}/\textbf{15.85}/\textbf{24.07}\\
\bottomrule
    \end{tabular}}
    \caption{Detailed ROUGE scores (R-1/R-2/R-L) on all the datasets in the few-shot setting (corresponds to Figure~\ref{fig:fewshot})}
    \label{tab:rouge_score_fewshot}
\end{table}

\section{Detailed Analysis on Fully Supervised Experiments}
\label{sec:detailed_fully_supervised}

To show the advantage of our pre-trained model when there is sufficient data, we also train the model with the full training set, and the results can be found in Table~\ref{tab:multinews_full_supervisory}-\ref{tab:arxiv_full_supervisory}\footnote{Due to the lack of computational resources, we do not train the model on Wikisum.}, along with the results from previous works. 
Differently from the zero-/few-shot experiments, here we report  the state-of-the-art results on different datasets, as they were presented  in the corresponding original papers. Since we use the same train/valid/test set as in those prior works, we can perform a fair comparison , without re-running all those extremely time-consuming experiments .

Overall, our model achieves state-of-the-art on Multi-News (see Table~\ref{tab:multinews_full_supervisory} 
, WCEP dataset (see Table~\ref{tab:wcep_full_supervisory}) and arXiv dataset (see Table~\ref{tab:arxiv_full_supervisory}).

\begin{table}[h!]
    \centering
    \resizebox{\linewidth}{!}{
    \begin{tabular}{l|c|c|c}
     Models    & ROUGE-1 & ROUGE-2 & ROUGE-L \\
     \hline
     
     \pegasus~\cite{pegasus}&47.52&18.72&24.91  \\
     BART-Long-Graph~\cite{pasunuru-etal-2021-efficiently} &49.03	&19.04 & 24.04 \\
     BART-Long-Graph(1000)~\cite{pasunuru-etal-2021-efficiently} & 49.24&18.99&23.97\\
        BART-Long(1000)~\cite{pasunuru-etal-2021-efficiently}&49.15&19.50&24.47\\
        \hhline{====}
        Ours & \textbf{49.94}&\textbf{21.05}&\textbf{25.85}\\
        \hline
    \end{tabular}}
    \caption{ROUGE scores of the previous models and our fully supervised model on the Multi-News dataset. The results of \pegasus is from \citet{pegasus}, and the other results are from \citet{pasunuru-etal-2021-efficiently} 
    }
    \label{tab:multinews_full_supervisory}
\end{table}

\paragraph{Multi-News} 
The experiment results on Multi-News dataset can be found in Table~\ref{tab:multinews_full_supervisory}.
Specifically, the \pegasus model~\cite{pegasus} is pre-trained on a large-scale single-document dataset with the Gap Sentence Generation objective, which is the same as ours, but with a different masking strategy, BART-Long~\cite{pasunuru-etal-2021-efficiently} uses the same model structure as ours 
, and BART-Long-Graph~\cite{pasunuru-etal-2021-efficiently} additionally has discourse graph injected. Comparing the results with the BART-Long model, our model is around 1 ROUGE point higher, which may result from either better model structure or pre-training. Interestingly, in one of the ablation studies in \citet{pasunuru-etal-2021-efficiently}, they find that the BART-Long model achieves its best performance with the length limit of 1000, and no further improvement is found when the length limit is greater than that. Thus we may conclude the gap between the performances is mainly from our design on the model, i.e. the document separators, proper global attention as well as the pre-training on a multi-document dataset.

\begin{table}[h!]
    \centering
    \resizebox{\linewidth}{!}{
    \begin{tabular}{lrrr}
    \toprule
     Models    & R1 & R2 & RL* \\
     \midrule
     LEAD&27.46&4.57&-\\
     BERTABS&31.56&5.02&-\\
BART&32.83&6.36&-\\
SCIBERTABS&32.12&5.59&-\\
     SOTA(Pointer Generator)&\textbf{34.11}&6.76&18.2\\
        \hhline{====}
        LEAD(ours) &26.49&4.26&14.70\\
        Ours & 31.93&	\textbf{7.37}&	18.02\\
        \bottomrule
    \end{tabular}}
    \caption{ROUGE scores of the previous models and our fully supervised model on the Multi-Xscience dataset. All the results are from \citet{lu-etal-2020-multi-xscience}. * The ROUGE-L is not comparable as we have different settings on the settings of evaluation, see the gap between LEAD and LEAD(ours).}
    \label{tab:multixscience_full_supervisory}
\end{table}

\begin{table}[h!]
    \centering
    \setlength{\tabcolsep}{2pt}
    \resizebox{\linewidth}{!}{
    \begin{tabular}{lrrr}
    \toprule
     Models    & R1 & R2 & RL \\
     \midrule
     \textsc{BertReg} \cite{gholipour-ghalandari-etal-2020-large}&35.0 &13.5& 25.5\\
SUBMODULAR+ABS\cite{gholipour-ghalandari-etal-2020-large}&
30.6 &10.1 &21.4\\
     DynE~\cite{dyne}	&35.4&	15.1	&25.6\\
        \hhline{====}
        Ours & \textbf{46.08}&	\textbf{25.21}&	\textbf{37.86}\\
        \bottomrule
    \end{tabular}}
    \caption{ROUGE scores of the previous models and our fully supervised model on the WCEP dataset. 
    }
    \label{tab:wcep_full_supervisory}
\end{table}

\paragraph{WCEP} As for the WCEP dataset, \textsc{BertReg} \cite{gholipour-ghalandari-etal-2020-large} is a Regression-based sentence ranking system with BERT embedding, which is used as extractive summarization method, while Submodular+Abs is a simple two-step abstractive summarization model with a submodular-based extractive summarizer followed by a bottom-up abstractive summarizer~\cite{gehrmann-etal-2018-bottom}. DynE is a BART-based abstractive approach, which is to ensemble multiple input, allowing single document summarization models to be directly leveraged on the multi-document summarization task.
Our model outperforms all the models by a large margin, including the SOTA model DynE, and it may indicate that the plain structure is more effective than purely ensembling the output of single documents.

\begin{table}[h!]
    \centering
    \resizebox{\linewidth}{!}{
    \begin{tabular}{@{}lrrr@{}}
    \toprule
     Models    & R1 & R2 & RL \\
     \midrule
     \pegasus(1K)&44.21&16.95&38.83\\
     Bigbird-\pegasus (3k) &	46.63&19.02&41.77\\
LED(4K)&	44.40&17.94&39.76\\
LED(16K)&	46.63 &19.62& 41.83\\
        \hhline{====}
        Ours(4k) & \textbf{47.58}&\textbf{	20.75}&	\textbf{42.57}\\
        \bottomrule
    \end{tabular}}
    \caption{ROUGE scores of the previous models and our fully supervised model on the arXiv dataset. The result of \pegasus and BigBird-\pegasus are from~\cite{bigbird}, and the results of LED are from~\cite{longformer}. The number in the parenthesis indicates the length limit of the input.}
    \label{tab:arxiv_full_supervisory}
\end{table}
\paragraph{arXiv} In addition to the experiments on multi-document summarization datasets, we also compare our fully supervised model with previous works on the arXiv dataset, with each section treated as a single document. All the models to be compared with are based on pre-trained models, and Bigbird-\pegasus and LED utilize the pre-training of \pegasus~\cite{bigbird} and BART~\cite{lewis-etal-2020-bart}, respectively. However, both Bigbird and LED apply more efficient attentions, which make the models able to take longer input (3k for BigBird, 4K and 16k for LED). Our model has a better performance than all the models, including LED(16K), which allows for the input 4 times longer than ours. It is worth mentioning that LED(4K) has the same structure as our model, with the same length limit of the input, and with the pre-training on multi-document datasets, our model is more than 3 ROUGE point better than it, which shows that the strategy not only works for multi-document summarization but can also effectively improve single-document summarization for long documents.

\section{Examples of Generated Summaries}
We show an example (from Multi-News) of generated summaries by \sys and compared models trained with different number of examples in Table~\ref{tab:qual_example}. And we show an example from DUC2007 (which is one of the examples used for human evaluation) with generated summaries by \sys and two compared models in Table~\ref{tab:qual_example_duc2007}, with all the models trained on 10 data examples from DUC2007.

\section{Software and Licenses}
Our code is licensed under Apache License 2.0. Our framework dependencies are: 
\begin{itemize}
    \item HuggingFace Datasets\footnote{\url{https://github.com/huggingface/datasets/blob/master/LICENSE}}, Apache 2.0
    \item NLTK \footnote{\url{https://github.com/nltk/nltk}}, Apache 2.0
    \item Numpy\footnote{\url{https://github.com/numpy/numpy/blob/main/LICENSE.txt}}, BSD 3-Clause "New" or "Revised"
    \item Spacy\footnote{\url{https://github.com/explosion/spaCy/blob/master/LICENSE}}, MIT 
    \item Transformers\footnote{\url{https://github.com/huggingface/transformers/blob/master/LICENSE}}, Apache 2.0
    \item Pytorch\footnote{\url{https://github.com/pytorch/pytorch/blob/master/LICENSE}}, Misc
    \item Pytorch Lightning \footnote{\url{https://github.com/PyTorchLightning/pytorch-lightning/blob/master/LICENSE}},Apache 2.0
    \item Longformer\footnote{\url{https://github.com/allenai/longformer/blob/master/LICENSE}}, Apache 2.0
    \item ROUGE \footnote{\url{https://github.com/google-research/google-research/tree/master/rouge}}, Apache 2.0
\end{itemize}
\begin{table*}[tb]
    \centering
    \scriptsize
    \begin{tabular}{@{}l|p{0.8\linewidth}@{}}
    \toprule
    Model& Summaries\\
    \midrule
    \sys -0 & 

    BALTIMORE (AP) — The U.S. Coast Guard says it's looking for a man who has cost the service about \$500,000 after responding to nearly 30 of his fake distress calls.. The Coast Guard also said such false distress alerts detract from their ability to respond to actual alerts.Anyone with information about the caller is asked to contact the Coast Guard's regional command center at 410-576-2525 or email investigators at CGIS-Baltimore@uscg.mil..\\
    \midrule
    \sys -10& The U.S. Coast Guard says it's looking for a man who has cost the service about \$500,000 after responding to nearly 30 of his fake distress calls. In a press release published Friday, the Coast Guard says the 28 calls have originated from around the area of Annapolis, Maryland. Each call involved the same male voice and used an emergency radio channel. The two most recent calls were made on the night of July 21 and the early morning of July 22. The Coast Guard also says hoax calls distract rescuers from real emergencies, putting both the public and the responding crews at risk. "Calls like these not only put our crews at risk, but they put the lives of the public at risk," said Lt. Cmdr. Sara Wallace, who heads the response sector in Maryland. "Calls like these not only put our crews at risk, but they put the lives of the public at risk." Anyone with information about the caller is asked to contact the Coast Guard's regional command center at 410-576-2525 or email investigators at CGIS-Baltimore@uscg.mil.\\
    \midrule
    \sys -100& The US Coast Guard is looking for a man who has cost the service about \$500,000 after making 28 false distress calls from Maryland. The 28 calls have involved the same male voice and used an emergency radio channel, the AP reports. The Coast Guard says it has responded to nearly 30 of the fake calls and that the estimated cost of the responses is \$500,000. The caller has been making the calls since July 2014. He could be facing a six-year prison sentence, a \$10,000 civil fine, and a \$250,000 criminal fine, the AP reports. "A hoax call is a deadly and serious offense," says Lt. Cmdr. Sara Wallace, head of the response sector in Maryland. "Calls like these not only put our crews at risk, but they put the lives of the public at risk."\\
    \midrule
    \pegasus -100 &The US Coast Guard has responded to 28 false distress calls in two years—and it's a man who's been making the calls, not the waves. The unidentified caller has made the calls in Annapolis, Maryland, since July 2014, the Coast Guard reports. Each call involved the same male voice and used an emergency radio channel. The two most recent calls were made on the night of July 21 and early morning of July 22. The Coast Guard says the calls were made in the hope that the public would get into the water. Rather than responding to actual alerts, the calls were made to divert the attention of rescue crews from real emergencies, the AP reports. The Coast Guard says the cost of the responses to these false alarms is about \$500,000. The caller made 28 false distress alerts from Annapolis, Maryland, beginning in July 2014, according to the Coast Guard. The estimated cost of the responses to these false alarms is \$500,000. The Coast Guard says the calls were made in the hope that the public would get into the water. Rather than responding to actual alerts, the calls were made to divert the attention of rescue crews from real emergencies, the AP reports.\\
    \midrule
    Fully Supervised & The US Coast Guard says it's looking for a man who has cost the service about \$500,000 after responding to nearly 30 of his fake distress calls, the AP reports. In a press release published Friday, the Coast Guard says the 28 calls have originated from around the area of Annapolis, Maryland. Each call involved the same male voice and used an emergency radio channel. The two most recent calls were made on the night of July 21 and the early morning of July 22. The Coast Guard also says hoax calls distract rescuers from real emergencies, putting both the public and the responding crews at risk. "A hoax call is a deadly and serious offense," says Lt. Cmdr. Sara Wallace, who heads the response sector in Maryland. "Calls like these not only put our crews at risk, but they put the lives of the public at risk." Making false distress calls is a felony, punishable by a maximum of six years in prison, a \$10,000 civil fine, \$250,000 criminal fine, and a reimbursement to the Coast Guard for their efforts, the Coast Guard says.\\
    \midrule
    Ground Truth & The US Coast Guard says it's looking for a man who has cost the service about \$500,000 after responding to nearly 30 of his fake distress calls, reports the AP. In a press release published Friday, the Coast Guard says the 28 calls have originated from around the area of Annapolis, Maryland. Each call involved the same male voice and used an emergency radio channel. He's been making the calls since July 2014. The two most recent calls were made on the night of July 21 and the early morning of July 22. “A hoax call is a deadly and serious offense,” a Coast Guard rep tells NBC4, which notes that such calls are a felony that carry six years in prison, \$10,000 civil fine, \$250,000 criminal fine, and reimbursement to the Coast Guard. "Calls like these not only put our crews at risk, but they put the lives of the public at risk.”\\
    \bottomrule
    \end{tabular}
    \caption{Generated summaries from \sys and best baseline model (according ROUGE score on this example) trained with different number of training examples. The data used here is the \#10 in the test set of Multi-News dataset on Huggingface.}
    \label{tab:qual_example}
    \end{table*}

\begin{table*}[tb]
    \centering
    \scriptsize
    \begin{tabular}{@{}l|p{0.8\linewidth}@{}}
    \toprule
    Model& Summaries\\
    \midrule
    \pegasus& In 1996, Congress passed the Line-Item Veto Act, which gave the president the power to cut individual projects from tax and spending bills without vetoing the entire legislation. The act was followed by the President's line-item veto, which he used to trim 144 million dollars from a 248 billion dollars defense spending bill. He also used the veto power to block a congressional rejection of his line-item veto on 38 military construction projects. The bill was passed by the House and the President signed it into law. The veto was challenged by members of both parties who said it was unconstitutional because it gave the president unchecked power to rewrite legislation. The Supreme Court agreed on Friday to hear argument and decide the constitutionality of the president line-item veto. In 1998 the President used his line-item veto to cut \$38 million from a military construction bill. In 1999 the President used his line-item veto to cut \$54 million from a military spending bill. In 2000 the President used his line-item veto to cut \$54 million from a defense spending bill. In January the President vetoed a tax and spending bill, which gave him the power to cut individual projects from tax and spending bills without vetoing the entire legislation. In February the President vetoed a spending bill, which gave him the power to cut individual projects from tax and spending bills without vetoing the entire legislation. In September the President used his line-item veto to cut \$54 million from a defense spending bill. The bill was rejected by the House and the President vetoed it. In November the President used his line-item veto to trim 144 million dollars from a defense spending bill.\\
    \midrule
    LED&In 1996, the Republican-led Congress passed the Line Item Veto Act, giving the president the power to delete individual items of spending and tax bills. Clinton used the power to cut individual projects from tax and spending bills. In February 1999, the President Clinton vetoed a congressional rejection of his line-item veto on 38 military construction projects. In May 1999, the President Clinton used the line-item veto to cut individual items of spending and tax breaks. In 2000, the President Clinton used the line-item veto to cancel individual items of spending and tax breaks. In May 2000, the President Clinton threatened to use the line-item veto to cancel all military spending and tax breaks. In June 2000, the President Clinton used the line-item veto to cut individual items of spending and tax breaks. In August 2000, the President Clinton used the line-item veto to cut individual items of spending and tax breaks. In September 2000, the President Clinton used the line-item veto to cut individual items of spending and tax breaks. In 2000, the President Clinton used the line-item veto to cut individual items of spending and tax breaks. In 2001, the President Clinton used the line-item veto to cut individual items of spending and tax breaks. In June 2000, the President Clinton used the line-item veto to cut individual items of spending and tax breaks. In August 2000, the President Clinton used the line-item veto to cut individual items of spending and tax breaks. In September 2000, the President Clinton used the line-item veto to cut individual items of spending and tax breaks. In 2000, the President Clinton used the line-item veto to cancel individual items of spending and tax breaks. In 2001, the President Clinton used the line-item veto to cut individual items of spending and tax breaks. In June 2000, the President Clinton used the line-item veto to cut individual items of spending and tax breaks. \\
    \midrule
    \sys & In 1996, Congress gave the President the power to veto specific items of tax and spending bills. Before the law's enactment in 1996, the president could veto an entire spending measure but not individual items. The court ruled that such a specialized veto can be authorized only through a constitutional amendment. In January 1997, the line-item veto law was passed. It was passed under the Republican Party's "Contract with Congress". It was passed after President Clinton vetoed thirteen relatively obscure research and spending programs, almost all of the military spending increases approved by Congress. In October 1998, Clinton used his line-item veto authority to have trimmed 144 million U.S. dollars from a 248 billion defense spending bill. In November 1998, Clinton vetoed 38 military construction projects, worth 287 million U.S. dollars. In February 1999, the Justice Department appealed the line-item veto law to the Supreme Court, which agreed to hear argument and decide the constitutionality of the law. Earlier this month, a federal judge struck down the line-item veto law as unconstitutional. The highest court's review will yield a momentous balance of powers ruling. The case is scheduled to be argued before the justices on April 27. The line item veto, strongly supported by President Bill Clinton and a number of his predecessors, was passed in 1996 under the Republican Party's "Contract with Congress". It was passed in January 1997. Before the law's enactment, the only way presidents could reject spending laws was to veto whole budget bills. In 1996, Congress gave the president the power to cancel individual items in tax and spending bills. In January 1997, the line-item veto law was passed. It was passed under the Republican Party's "Contract with Congress". It was passed in January 1997. In 1998, President Clinton threatened to veto some items of the military construction bill because of the increased funding. In November 1998, Clinton used his line-item veto power to delete 38 projects in 24 states worth 287 million U.S. dollars. In February 1999, the Justice Department appealed the line-item veto law to the Supreme Court, which agreed to hear a case about its constitutionality. \\
    \midrule
    Ground Truth & In 1996 a Republican congress overwhelmingly passed a Line Item Veto Act allowing presidents (including the incumbent Democratic president), to strike individual tax or spending items within 5 days after signing a bill into law. Congress could restore those items in a new bill passed by majority vote. If the president vetoed that bill, Congress could override that veto with a two-thirds majority. Proponents argued that the law preserved the integrity of federal spending, saved billions of dollars, and that it did not repeal any portion of a law, but was simply a delegated spending authorization from Congress. In January 1997, the first year of the law, the president vetoed 163 line-items in six bills, and in 1998 82 line-items in 11 bills. In October 1997 Congress overrode the president's line-item veto against 36 of 38 military construction projects. Initial 1997 efforts by congressmen to challenge the law in the Supreme Court were rejected due to lack of standing. On June 25, 1998 after lower courts rejected the Line Item Veto Act as unconstitutional, on appeal by the White House the Supreme Court ruled 6-3 that Congress unconstitutionally violated the principle of separation of powers, because that procedure allows the president to create a law that was not voted on by either house of Congress in violation of the Constitution's Article I "presentment" clause. A constitutional amendment would be required to institute line item vetoes. Justices Breyer and Scalia argued similar dissenting opinions that separation of powers was not violated.\\

    \bottomrule
    \end{tabular}
    \caption{Generated summaries from \sys, \pegasus and LED trained with 10 training examples, along with one (out of four) ground-truth summary. The data used here is D0730 in DUC2007.}
    \label{tab:qual_example_duc2007}
    \end{table*}

\section{Annotation Instructions for Human Evaluation}
\begin{figure*}
    \centering
    \includegraphics[scale=0.8]{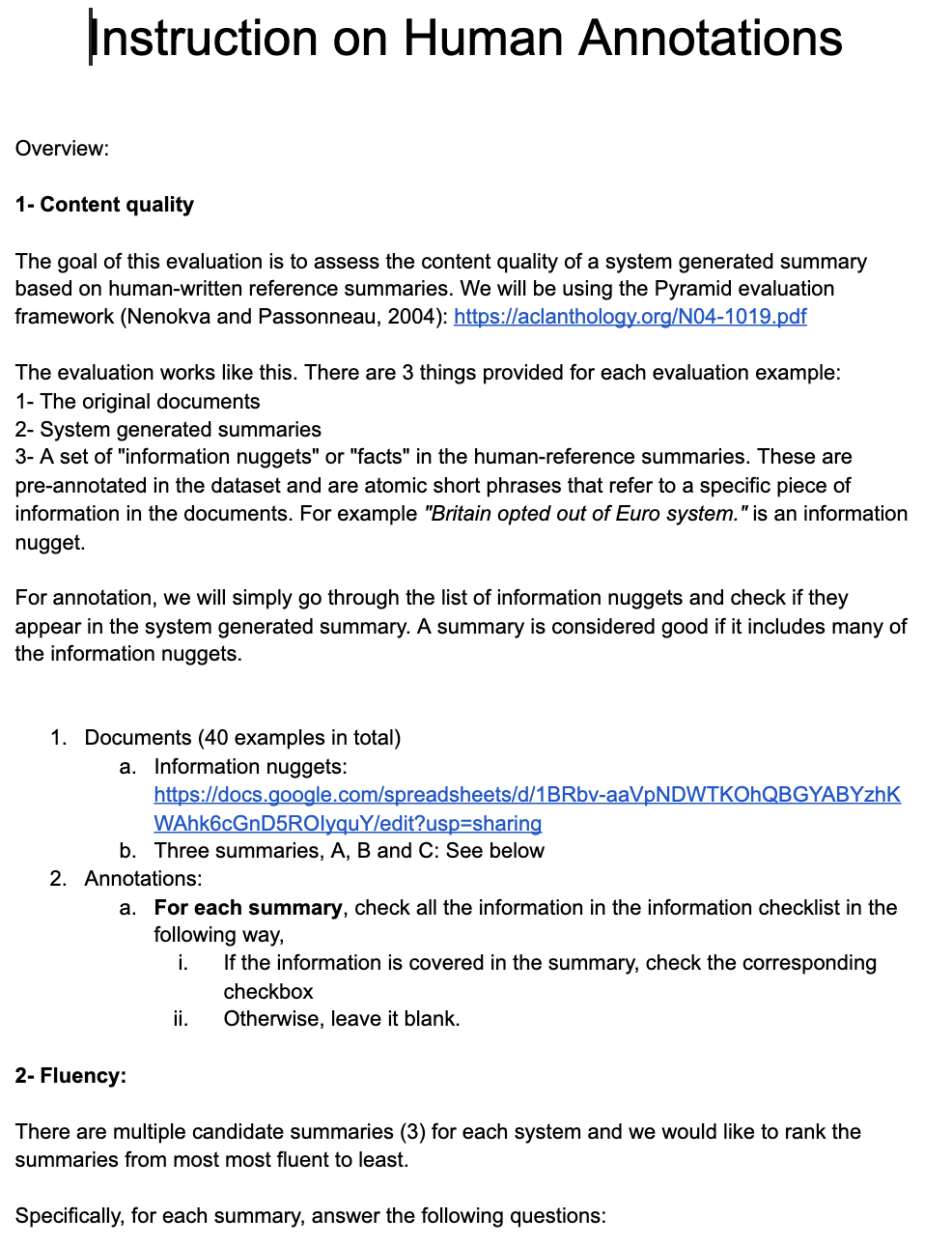}
    \caption{Annotation instruction for human annotators.}
    \label{fig:humaneval_1}
\end{figure*}

\begin{figure*}
    \centering
    \includegraphics[scale=0.8]{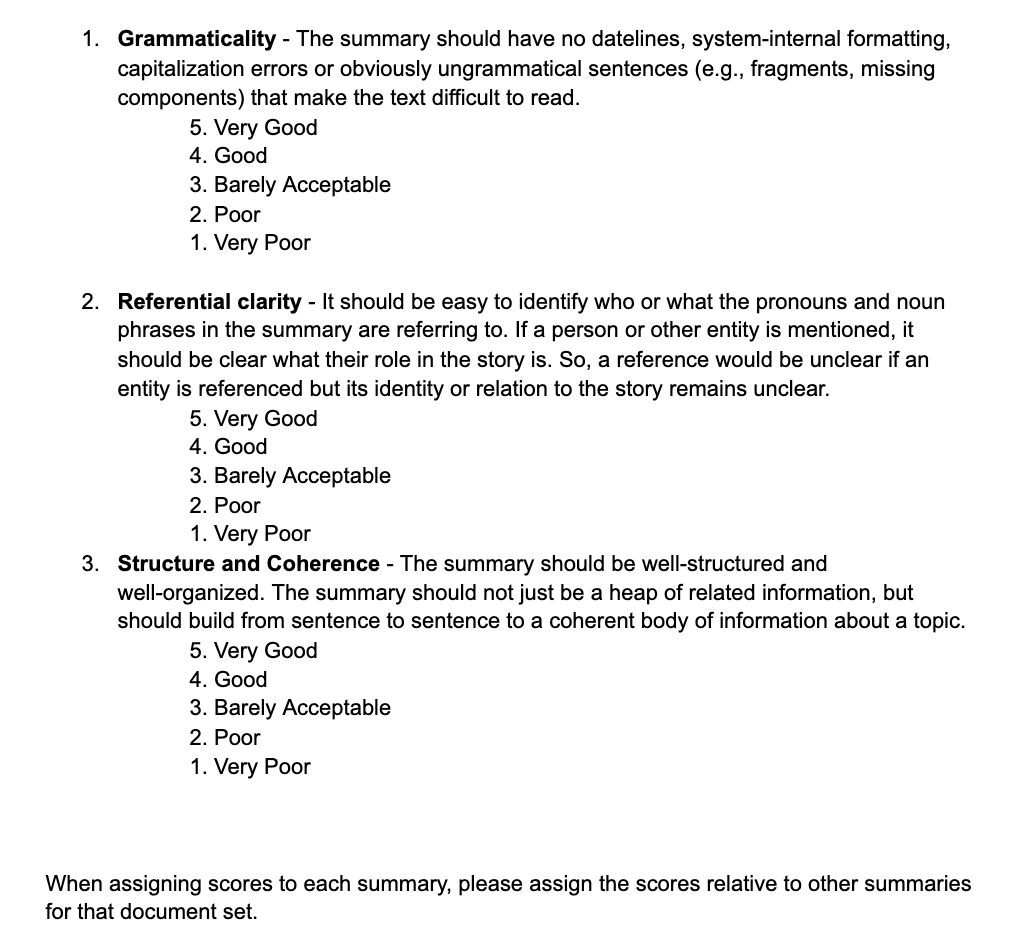}
    \caption{Annotation instruction for human annotators.}
    \label{fig:humaneval_2}
\end{figure*}

Figure~\ref{fig:humaneval_1} and Figure~\ref{fig:humaneval_2} shows the annotation instruction for human annotators.

\end{document}